# A Sequential Thinning Algorithm For Multi-Dimensional Binary Patterns

Himanshu Jain, Archana Praveen Kumar

*Abstract*—Thinning is the removal of contour pixels/points of connected components in an image to produce their skeleton with retained connectivity and structural properties. The output requirements of a thinning procedure often vary with application. This paper proposes a sequential algorithm that is very easy to understand and modify based on application to perform the thinning of multi-dimensional binary patterns. The algorithm was tested on 2D and 3D patterns and showed very good results. Moreover, comparisons were also made with two of the state-of-the-art methods used for 2D patterns. The results obtained prove the validity of the procedure.

## I. INTRODUCTION

Thinning of a binary pattern, also referred to as skeletonization is the deletion of outline pixels of a connected component while maintaining their basic structure and connectivity. This helps in reduction in the amount of foreground data for further processing and also in shape analysis using feature extraction etc. The thinning algorithms in existence vary in their methodology greatly due to its complicated nature and produce different results for different patterns. The output requirements of a skeletonization algorithm also vary with application. For a general case however, a good algorithm is expected to satisfy the following criteria:

i) The resultant skeleton should be 1 pixel-wide with no redundant pixels.

ii) connectivity of components must be preserved,

iii) Excessive erosion/deletion of points must be avoided.

iv) The skeleton should be centered inside the component. (Fig. 1).

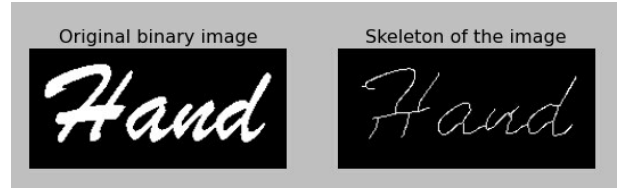

Figure 1: A simple skeletonization example. (Left: input image with 2 connected components, Right: Output skeleton image with 2 connected components)

Thinning methodologies are broadly classified into 2, iterative and non-iterative. Iterative methods repeat a procedure until the skeleton is obtained. Iterative methods may further be divided into sequential and parallel algorithms. Sequential algorithms are those in which the decision on a pixel is affected by the previously processed pixels within the same iteration. Parallel algorithms are those that depend only on the result obtained from the previous iteration. Non-Iterative methods usually try to extract the skeleton in a single pass.

## II. RELATED WORK

Most, if not all, thinning algorithms proposed till the present date were devised to handle patterns with a pre-specified number of dimensions. This paper on the other hand proposes a generic algorithm for all patterns having 2 dimensions or more. Thinning algorithms however see the most use in the thinning of 2D patterns. Therefore, 2 of the state-of-the-art and most widely used methods for 2D patterns have been discussed in this section. These algorithms were proposed by i) Zhang and Suen [1] and ii) Guo and Hall [2]. A qualitative and quantitative comparison of these methods against the proposed method has been made in section IV.

i) The ZS algorithm described in [1] is a parallel iterative algorithm with two sub-cycles in each iteration. The algorithm examines the 3x3 local neighborhood of a pixel (Figure 2) to decide if a pixel should be deleted. The first sub-cycle is used to delete southeast pixels. It deletes a pixel if the following 4 conditions are satisfied:

- $2 \leq BP \leq 6$
- $AP = 1$
- $P_2 \vee P_4 \vee P_6 = 0$
- $P_4 \vee P_6 \vee P_8 = 0$

Where BP is the number of foreground pixels in the 3x3 neighborhood of examined pixel, AP is the number of transitions from 0 to 1 in the ordered list of neighbors $P_2P_3P_4P_5P_6P_7P_8P_9P_2$.

| $P_9 = [x-1, y-1]$ | $P_2 = [x-1, y]$ | $P_3 = [x-1, y+1]$ |
| --- | --- | --- |
| $P_8 = [x, y-1]$ | $P_1 = [x, y]$ | $P_4 = [x, y+1]$ |
| $P_7 = [x+1, y-1]$ | $P_6 = [x+1, y]$ | $P_5 = [x+1, y+1]$ |

*Figure 2: 3X3 Local-neighborhood for the ZS algorithm*

In the second sub-iteration, north-west pixels are removed which satisfy the following 4 conditions:

- $2 \leq BP \leq 6$
- $AP = 1$
- $P_2 \vee P_4 \vee P_8 = 0$
- $P_2 \vee P_6 \vee P_8 = 0$

The border pixels which satisfy the criterion of first sub-iteration are marked for deletion. At the end of first sub-iteration, all the marked pixels are deleted. The same is then, repeated for the second sub-cycle. This finishes current iteration and these iterations are repeated until no pixel is deleted in either of the sub iterations.

This algorithm thins the digital patterns to one-pixel width and executes fast. But it suffers with a few drawbacks. First, it is insensitive to small noise near the north-east and south-west corners. Second, it doesn't preserve 8-connectivity in all cases. Besides this, it completely eliminates patterns consisting of 2x2 squares. Another drawback is that the algorithm may leave behind some 2 -pixel wide diagonal lines in the skeleton.

ii) Guo and Hall [2] proposed a two sub-iteration parallel thinning algorithm and compared it with ZS algorithm. To preserve the end points and at the same time to remove redundant pixels they proposed a new operator named NP = min (NP1, NP2). Where NP1 and NP2 are defined as follows:

$NP1 = (P_9 \vee P_2) \wedge (P_3 \vee P_4) \wedge (P_5 \vee P_6) \wedge (P_7 \vee P_8)$

$NP2 = (P_2 \vee P_3) \wedge (P_4 \vee P_5) \wedge (P_6 \vee P_7) \wedge (P_8 \vee P_9)$

Thus, the algorithm deletes pixels if the following conditions are satisfied:

i) CP = 1, where CP = number of connected components in the 3x3 neighborhood of pixel.
ii) NP = 2 **or** NP = 3
iii) (a) $(P_2 \vee P_3 \vee NOT(P_5)) \wedge P_4$ (for odd numbered iteration) or
  (b) $(P_6 \vee P_7 \text{ OR } NOT(P_9)) \wedge P_8$ (for even numbered iteration)

Condition 1 preserves 8-connectivity. Criteria (iii-a) deletes north-east pixels whereas (iii-b) removes south-west pixels.

By deleting redundant pixels, this algorithm produces thinner skeletons compared to the ZS algorithm but does not at times retain the structure of the component accurately. It produces medial axis points for horizontal, vertical and diagonal lines. It also produces the final skeleton in fewer iterations.

### III. METHODOLOGY

The proposed method uses a sequential iterative technique similar to raster scan to obtain the thinned pattern. The number of sub-cycles required within each iteration is equal to the number of dimensions of the pattern k, and the iterations are repeated until no more changes occur. The detailed procedure to perform the thinning is described below.

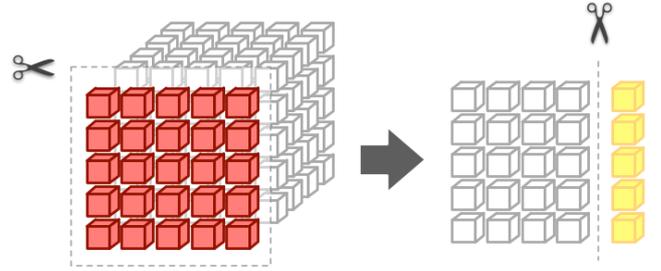

*Figure 3: 3D Slicing (Yellow pixels represent one slice in the vertical dimension)*

Assume a multi-dimensional pattern of size $N_1 \times N_2 \times N_3 \ldots N_k$ where $N_i$ is the size of the ith dimension. Now, for every dimension i, we extract the set of $(N_1 * N_2 * N_3 \ldots N_k) / N_i$ slices of size $1 \times N_i$. The slicing of a 3D pattern is depicted in fig. 3. For a 2D pattern, these slices would simply be the set of rows in the 2D array for the horizontal dimension and columns for the vertical dimension. The extracted slices are then, processed one after the other starting from the first slice as described below.

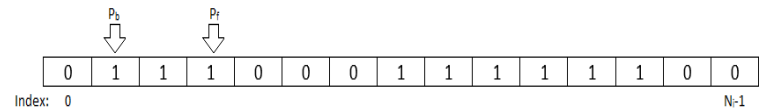

*Figure 4: $P_b$ and $P_f$ for a slice of size $1 \times N_i$*

For each slice, we extract the set of connected pixels/points in the slice. For each set of connected pixels, define a forward pixel ($P_f$) and a backward pixel ($P_b$) as shown in fig.4. The forward pixel is the pixel with the highest index in the ith dimension and the backward pixel is the pixel with the lowest index in the ith dimension. These indicate the contours of the connected components along that slice. If $P_f$ is the same as $P_b$, the pattern is already of 1-pixel width in that slice and no further operations need to be performed. If it is not, we examine the $3^k$ local-neighborhoods of $P_f$, $P_b$ to decide if they may be deleted. For a 2D image this local neighborhood would be of size 3 X 3 as shown in fig. 2 and for a 3D image, it would be 3 X 3 X 3. The neighborhoods are examined as follows:

i) Forward pixel ($P_f$):

For the forward pixel, we examine the set of neighbors in the $3^k$ local neighborhood. If the total number of foreground pixels in the neighborhood is equal to 2 (including $P_f$), the pixel is an end-point and must be retained. If it is not, we look at the foreground pixels in the neighbors that are ahead of $P_f$, i.e. whose index in the $i^{th}$ dimension is greater than that of $P_f$. For each of these foreground pixels F, we extract their $3^k$ neighborhood. From this neighborhood, we take into account the subset of the neighbors that are behind F, i.e. whose index in the $i^{th}$ dimension is lesser than that of F. We then take the intersection of these neighbors with $P_f$'s neighborhood as in (1). If all the pixels in this intersection S are background pixels, $P_f$ must be retained as it is important to the connectivity of the component. Else, the pixel may be deleted/eroded.

$$S = N_F \cap N_P - P \quad (1)$$

where $N_P$ is the $3^k$ neighborhood of the forward/backward pixel P, and $N_F$ is the sub-set of neighbors in the $3^k$ neighborhood of the foreground pixel F.

ii) Backward Pixel ($P_b$):

The backward pixel is examined in a manner similar to the foreground pixel, as follows:

For the forward pixel, we examine the set of neighbors in the $3^k$ local neighborhood. If the total number of foreground pixels in the neighborhood is equal to 2 (including $P_b$), the pixel is an end-point and must be retained. If it is not, we look at the foreground pixels in the neighbors that are behind $P_b$, i.e. whose index in the $i^{th}$ dimension is lesser than that of $P_b$. For each of these foreground pixels F, we extract its $3^k$ neighborhood. From this neighborhood, we take into account the subset of the neighbors that are ahead of F, i.e. whose index in the $i^{th}$ dimension is greater than that of F. We then take the intersection of these neighbors with $P_b$'s neighborhood as in (1). If all the pixels in this intersection set are background pixels, $P_b$ must be retained as it is important to the connectivity of the component. Else, the pixel may be deleted/eroded. A trivial condition, however, must be checked before deletion of the backward pixel is performed. This trivial condition is to ensure that the pixel immediately ahead of $P_f$ is not a background pixel. Such a scenario would indicate that the pattern has already achieved 1-pixel width along that slice. (This is possible when $P_f$ and $P_b$ are 2 consecutive pixels in the $i^{th}$ dimension and $P_f$ was already deleted.)

The iterations with k sub-iterations are repeated until no more changes occur in the pattern. This concludes the thinning procedure. The pseudocode for thinning in 2 dimensions is given in table 1 for further understanding.

IV. EXPERIMENTAL RESULTS AND DISCUSSION

Experiments were performed on patterns of i) 2 dimensions and ii) 3 dimensions. The skeletons obtained in both types of patterns were well acceptable and retained the structural properties of the components while bringing the thickness down to 1-pixel width as much as possible. From the results it was also observed that the algorithm is very insensitive to noise. Such noise can be handled by either preprocessing the binary pattern or by postprocessing the obtained skeleton using procedures like smoothening. The algorithm may be also modified by the user in cases where the nature of the noise is known. The detailed results for the 2 types of patterns are discussed in subsections IV(A) and IV(B).

A. 2D patterns

Experiments were performed on 2 sets of 2D patterns, i) the IAM database [21] which consists of images of handwritten and digitally printed text, since thinning is often used in the context of OCR, ii) a set of regular shapes to understand the working and modifiability of the algorithm. The results of the skeletons obtained were further compared with the algorithms discussed in section II.

The quantitative comparison of the results of the proposed algorithm against the ZS and GH algorithms was performed using some of the evaluation measures described by Jang and Chin in [3]. These measures are described as follows:

i) Measure of Convergence to Unit Width

This indicates how thin or how close to 1-pixel width the obtained skeleton is. It is defined as,

$$m_t = 1 - \left(\frac{Area[\cup_{1 \leq k \leq 4} S_m Q^k]}{Area[S_m]}\right)$$

where Area[.] is the operation that counts the number of foreground pixels, $S_m$ is the resultant skeleton, and Q is the set of structuring elements (fig. 5) used for hit or miss transform.

The value of $m_t$ is a non-negative integer in the range [0,1]. The greater this value the better is the convergence.

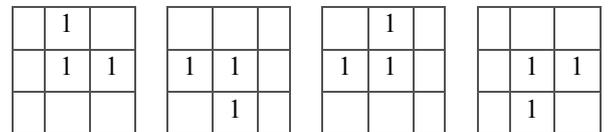

Figure 5: Set of structuring elements Q= {$Q^1$, $Q^2$, $Q^3$, $Q^4$}

ii) Size ratio

This is the ratio of the number of foreground pixels in the skeleton to the number of foreground pixels in the input binary image. This measure is significant because it gives an idea of the computation time needed for further processing. However, it must be considered only if the quality of the skeleton is acceptable. It is defined as,

$$D_r = \frac{|S_m|}{|B|}$$

where $S_m$ is the resultant skeleton, B is the input binary image, |x| is the operation that counts the number of foreground pixels in the pattern x. The lower this measure, the greater the reduction in foreground data.

*iii) Number of iterations (n)*

This measure is the total number of iterations undergone by the procedure until the result is obtained. Although this measure does not give an absolute measure of performance; a low value indicates that the method is robust. Execution times are not considered here since they vary greatly across different implementations.

**Table 1: Pseudocode for 2D patterns**

```
// returns the locations of neighbors in the 3 x 3 neighborhood of pixel at location (x,y)
function nbr_locs ( pixel , pattern ) :
        x , y = pixel
        return [ ( x – 1 , y – 1 ) , ( x – 1 , y ) , ( x – 1 , y + 1 ) ,
                ( x , y – 1 ) , ( x , y ) , ( x , y + 1 ) ,
                ( x + 1 , y – 1 ) , ( x + 1 , y ) , ( x + 1 , y + 1 ) ]
```

```
// Thinning function where pattern is a 2D array of size N₁ X N₂
// returns the skeleton
function thin (pattern) :
        while True :
                img2 ← img . copy()
                //delete along horizontal axis
                For x in range (0, pattern .shape [ 0 ] ) do : // pattern . shape= [N₁, N₂]
                        y ← 0
                        while y < pattern . shape [ 1 ] :
                                // assuming foreground pixels = 1 and background pixels = 0
                                if img [ x , y ] = 1 :
                                        P_b ← [ x , y ]
                                        while img [ x , y ] = 1 :
                                                y ← y + 1
                                        y ← y - 1 // backtrack one step
                                        P_f ← [ x , y ]
                                        if not ( P_f = P_b ) :
                                                //check for deletion of P_f
                                                N_p ← nbr_locs ( P_f, pattern )
                                                flag ← 0 // flag = 0 indicates that P_f is deletable

                                                //check for end-point
                                                if sum ( N_p ) <= 2 :
                                                        flag ← 1

                                                //check for connectivity
                                                for nbr in N_p do :
                                                        if nbr [ 1 ] > P_f[ 1 ] : // neighbor ahead of P_f
                                                                if pattern [ nbr [ 0 ] , nbr [ 1 ] ] = 1 :
                                                                        temp_nbrs = nbr_locs ( nbr , pattern )
                                                                        N_F ← [ ]
                                                                        for n in temp_nbrs do:
                                                                                if n [ 1 ] < nbr [ 1 ] :
                                                                                        N_f.add ( n )

                                                                        S ← N_f ∩ N_p – P_f
                                                                        if sum ( s ) = 0 :
                                                                                flag ← 1

                                                if flag = 0:
                                                        pattern [P_f[ 0 ] , P_f[ 1 ] ] ← 0

                                        if not ( pattern [ P_b [ 0 ] , P_b [ 1 ] – 1 ] ) = 0 : // trivial condition
                                                // check for deletion of N_b

                //delete along vertical axis

                if img2 = img :
                        break
        return img
```

The results of the experiments performed on the IAM database are tabulated in Table 2. The values listed in the table are average values from 529 different document images in the database. The proposed algorithm outperforms the ZS algorithm with respect to all the defined measures while handling its drawbacks. The algorithm also produces better values of $m_t$ and n than the Guo-Hall algorithm, however size ratio is marginally higher for the proposed algorithm. This indicates either excessive erosion of end-points in the Guo-Hall method or presence of spurious branches in the skeleton produced by the proposed method, both of which are unfavorable. Visual inspection of the skeletons indicates a combination of the two.

**Table 2: Quantitative measures for the IAM database**

| Algorithm | $S_r$ | $M_t$ | N |
|---|---|---|---|
| ZS | 0.564180451 | 0.86552364 | 6.23062381 |
| GH | 0.516420081 | 0.97771798 | 4.73345935 |
| Proposed | 0.517404482 | 0.98800664 | 4.09829867 |

The skeletons produced by the different algorithms for a few noisy samples in the database are shown in fig 5 and fig. 6. From the resultant skeletons it can be observed that the proposed algorithm does the best job at preserving the structure of the components (for example, the retention of the small branch on the top left corner in the letter 'n' as in fig. 5) but does so at the cost of insensitivity to noise (for example, the small inward branch in the letter 'o' as in fig. 6). From the point of view of OCR, signature verification etc., even the most minute features are important. In such cases especially, the proposed algorithm does a much better job at retaining features when compared to the other 2 methods. Moreover, the retention of end-points is also very high for the proposed method.

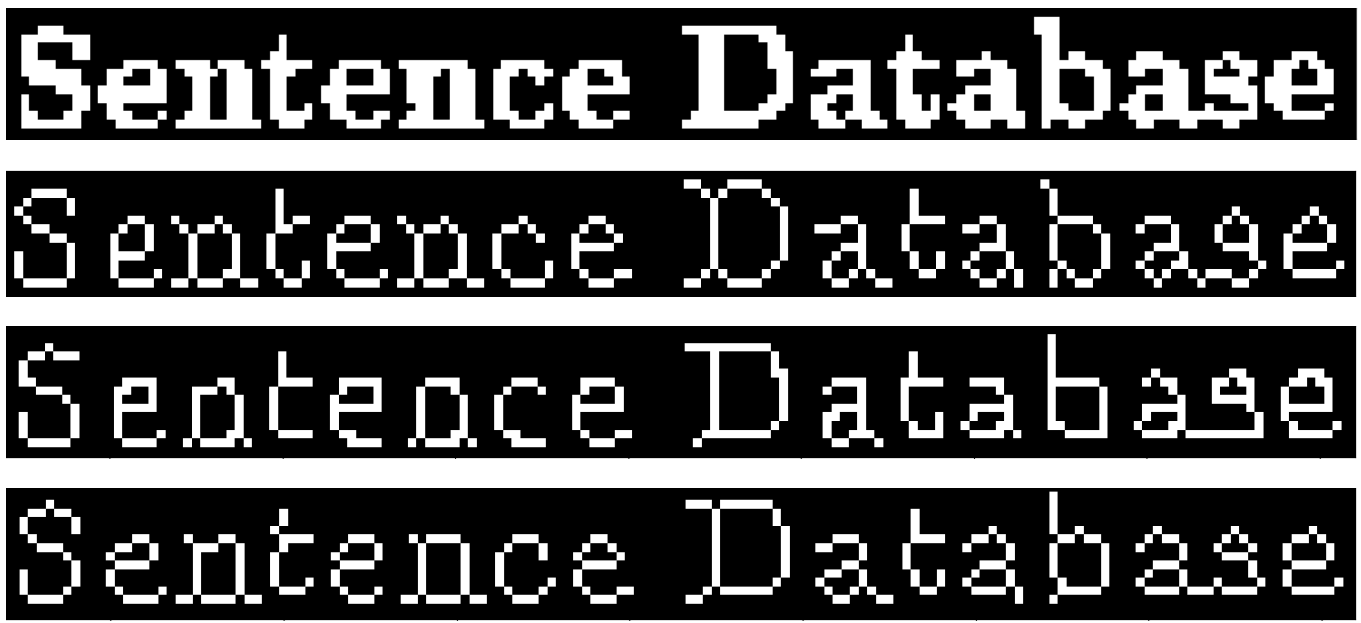

*Figure 5: Digitally printed text (from top to bottom: a) Original pattern, b) Guo-Hall algorithm, c) Zhang Suen algorithm, d) proposed algorithm)*

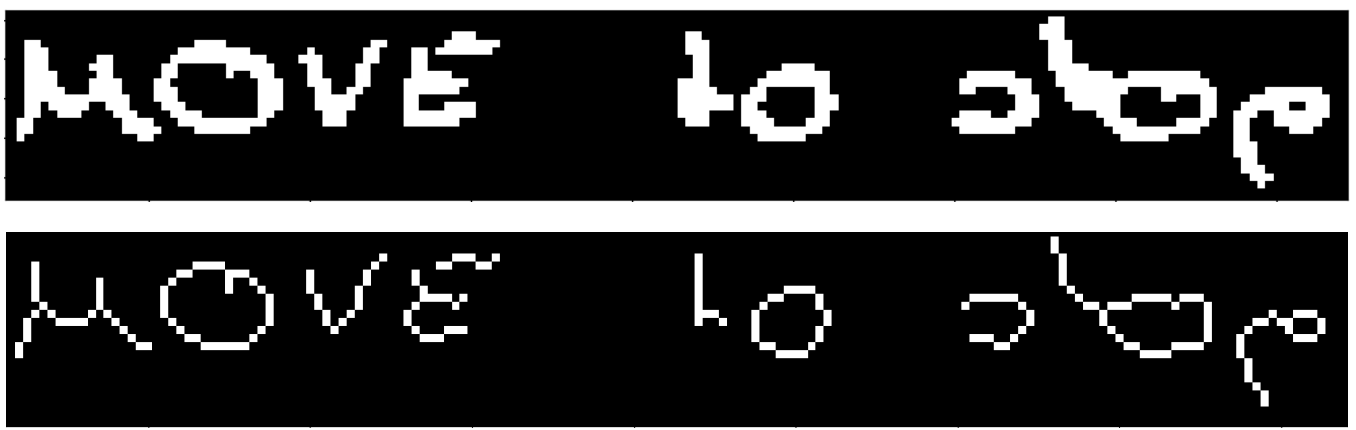

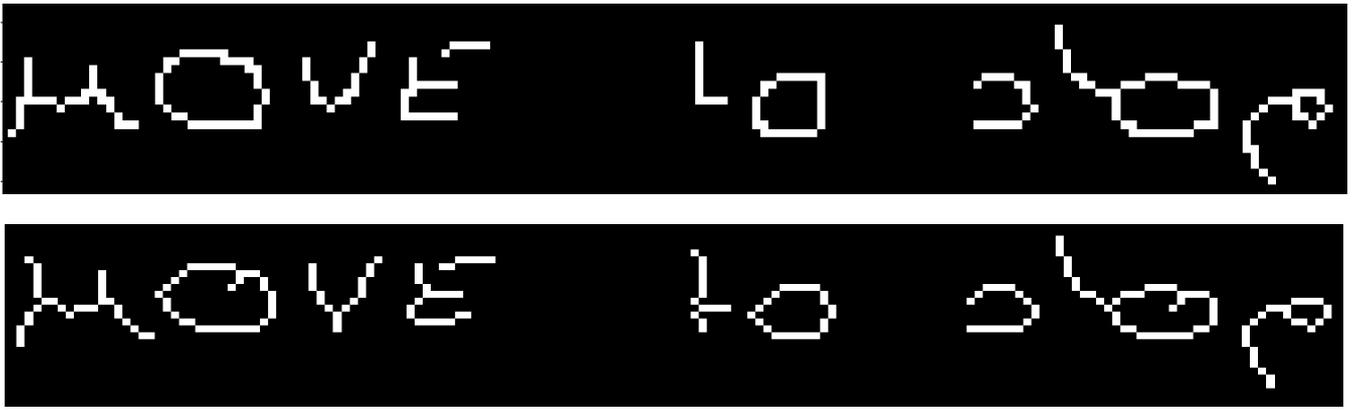

*Figure 6: Handwritten text (from top to bottom: a) original pattern, b) Guo-Hall algorithm, c) Zhang Suen algorithm, d) proposed algorithm*

Aside from the IAM database, the algorithms were also tested on a set of regular shapes like circles, triangles etc. to further understand and compare the quality and nature of the produced skeletons. The results are shown in table 3. The results obtained further demonstrate the ability of the proposed algorithm to preserve the original structure of the components (for example, the skeleton the lightning in table 3). Moreover, the proposed algorithm produces different skeletons for squares and circles while the other two do not. The proposed method, here therefore, possesses greater distinguishing ability.

**Table 3: Skeletons for simple shapes**

| Original Structure | GH algorithm | ZS algorithm | Proposed algorithm |
|---|---|---|---|
| 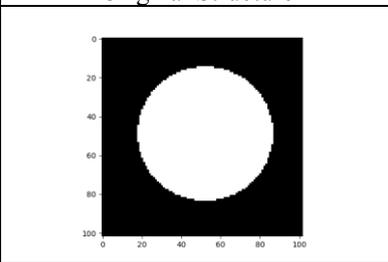 | 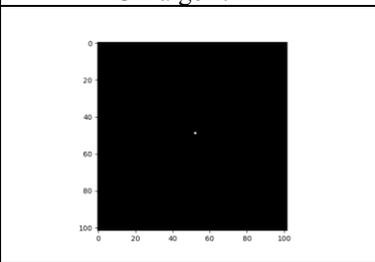 | 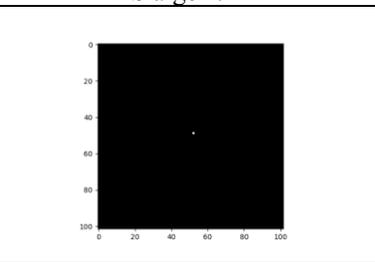 | 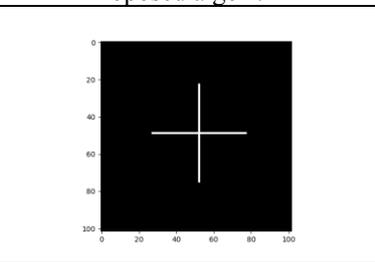 |
| 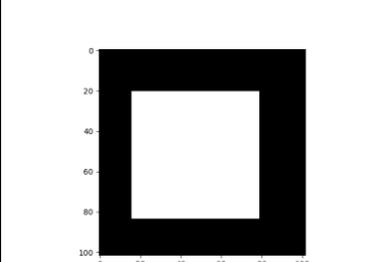 | 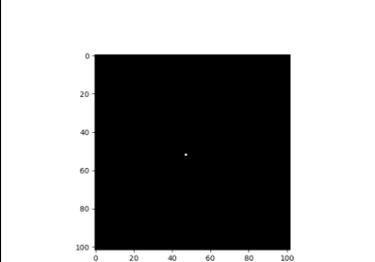 | 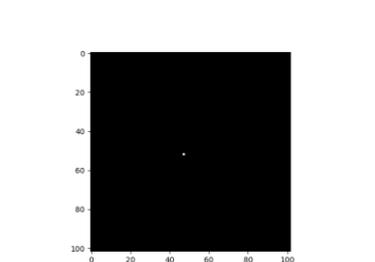 | 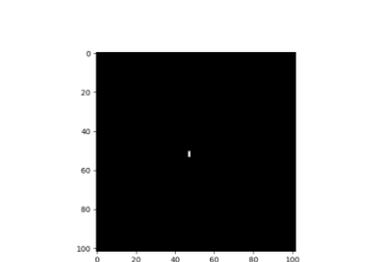 |
| 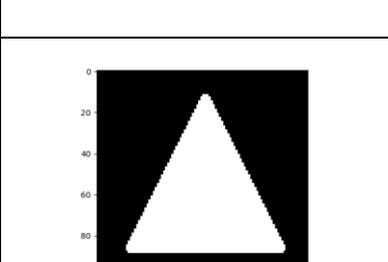 | 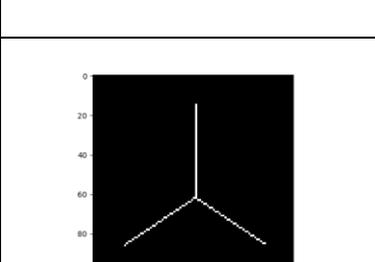 | 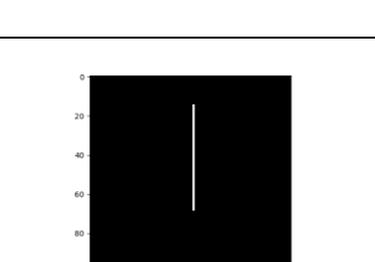 | 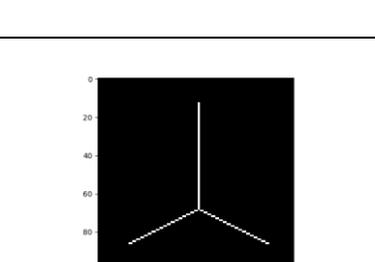 |

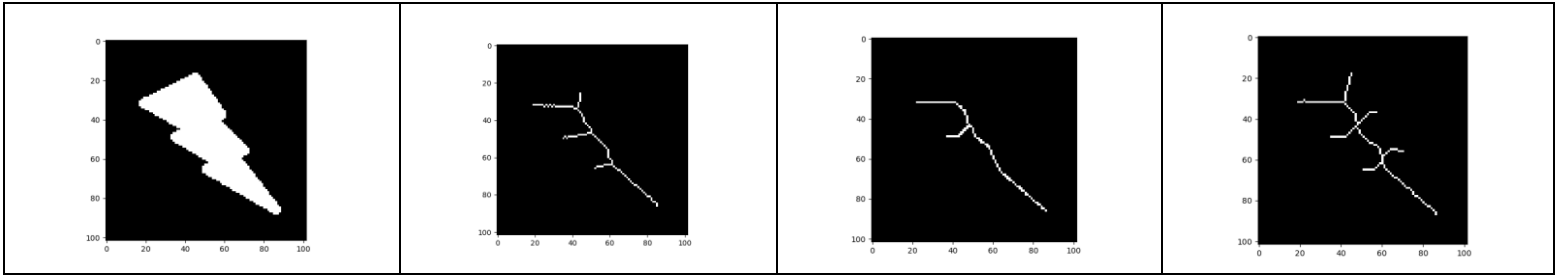

The highlight of the proposed method lies in the fact that it is very easy to understand and modify as per the requirement. It must be understood that, for 2D patterns, the deletion of forward and backward pixels as described in the methodology corresponds to deletion of the east and west contours of the connected components in the horizontal dimension and deletion of south and north contours in the vertical dimension. This understanding helps us modify the order of deletion, number of repetitions of each sub-cycle within each iteration etc. based on application. For example, it may be desirable for an application to retain the skeleton of a square as a vertical line in the center, while retaining its vertical length. This can be achieved by performing deletion of the contour pixels only along the horizontal dimension. A few such simple modifications are demonstrated in fig. 7.

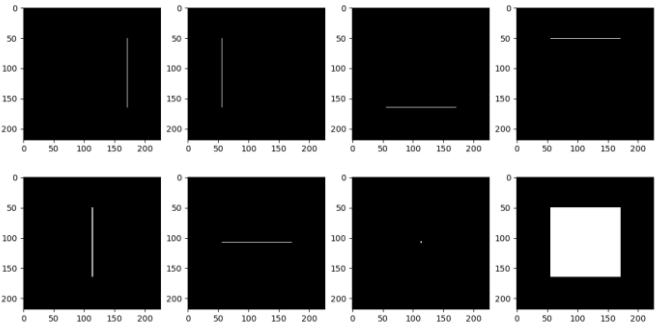

*Figure 7: Thinning of a square. (From top-left in clockwise order: i) west erosion (only backward pixel deletion in horizontal dimension), ii) only east erosion (only forward pixel deletion in the horizontal dimension) , iii) only north erosion (only backward pixel deletion in vertical dimension), iv) only south erosion ( only forward pixel deletion in vertical dimension), v) original image, vi) EWNS deletion (both vertical and horizontal deletion), vii) north-south deletion (only vertical dimension erosion), viii) east-west deletion (only horizontal dimension erosion)*

### B. 3D patterns

Experiments were performed on i) a set of common 3D volumetric objects like spheres etc. and ii) the 3D MNIST [22] database which consists of 3d patterns of numerical digits. No metrics were used however to quantify the results.

The results of performing the proposed procedure on the set of common 3D objects are depicted in table 4. It must be noted that any unexpected gaps and irregularities in the skeletons are due to limitations of the software used for visualization. Also, no surface/ axis in the resulting skeletons is more than 1-pixel wide unless otherwise expected. All figures were constructed symmetrical about the x-axis and y-axis. The algorithm can be modified similar to what was discussed for the square in section IV(A) to retain median planes etc. As can be seen the algorithm is highly sensitive to noise. For example, the cylinder for testing is rugged, which results in planes on the top and bottom when erosion is performed along the x-axis and y-axis. Generally, though, the skeletons obtained are good representations of their original patterns.

**Table 4: Skeletons of simple 3D objects**

| Shape | Original pattern surface | X-axis and Y-axis and Z-axis erosion | X-axis and Y-axis erosion | (Z-axis and X-axis) or (Z-axis and Y-axis) erosion | X-axis or Y-axis erosion | Z-axis erosion |
|---|---|---|---|---|---|---|
| One-sheet hyperbola (rugged) | | | | | | |
| Two-sheet hyperbola (rugged) | 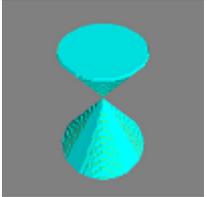 | 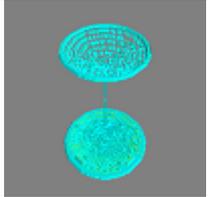 | 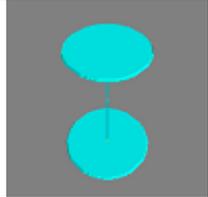 | 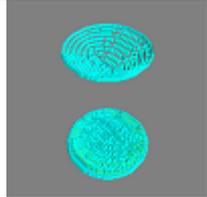 | 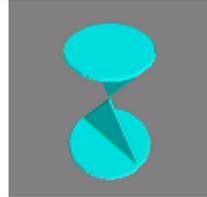 | 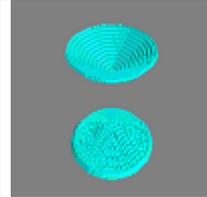 |
| Elliptic parabola (rugged) | 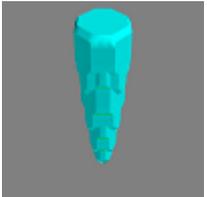 | 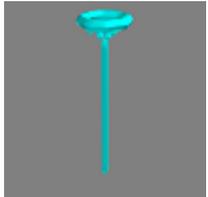 | 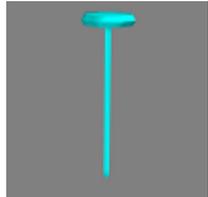 | 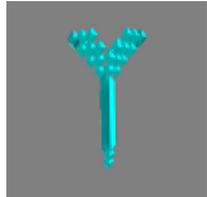 | 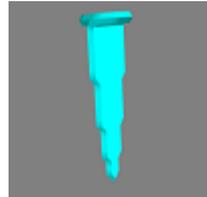 | 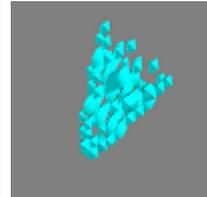 |
| Cylinder (rugged) | 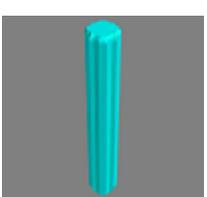 | 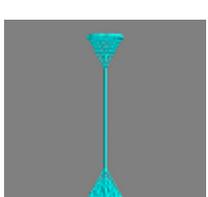 | 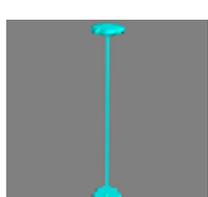 | 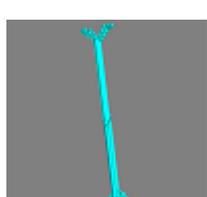 | 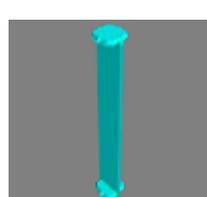 | 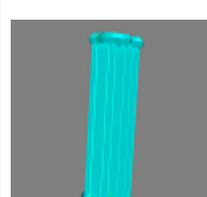 |
| Sphere (rugged) | 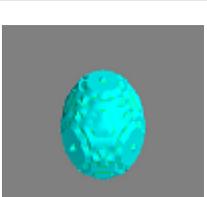 | 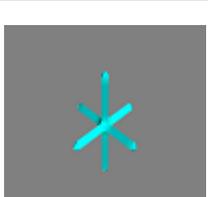 | 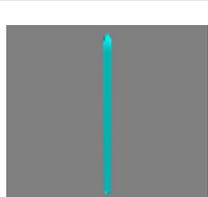 | 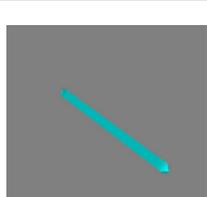 | 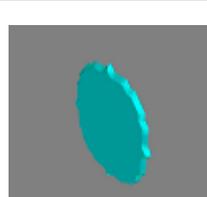 | 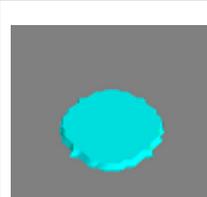 |

Aside from the regular 3D objects used to test the procedure, experiments were also performed on the MNIST database. The medial axis obtained via erosion of points along all the axis/dimensions, through good representations, hold little meaning and are hard to interpret. Instead, the 3D patterns of the objects can be converted to their 2D representations, i.e. medial plane representations which can then be thinned further to obtain the medial axis representations. Such representations, though not perfect medial axis, are much easier to interpret. This is one simple example of how the algorithm may be modified according to application. To do this, first the 3D patterns are thinned/eroded along only the z-axis and then alternatingly along the x-axis and y-axis. The results obtained for a few patterns in the 3D MNIST database are depicted in table 5.

**Table 5: 3D MNIST skeletons**

| Digit | Original pattern | Medial plane | Medial axis |
|---|---|---|---|
| 1 | 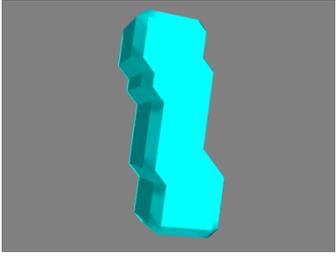 | 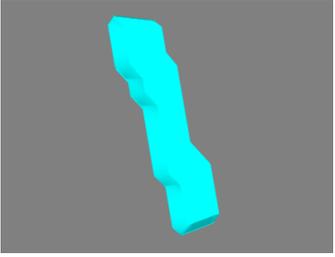 | 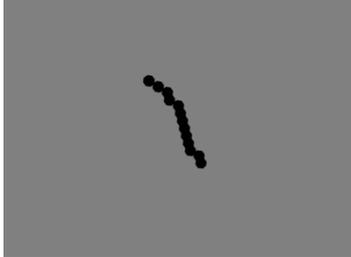 |

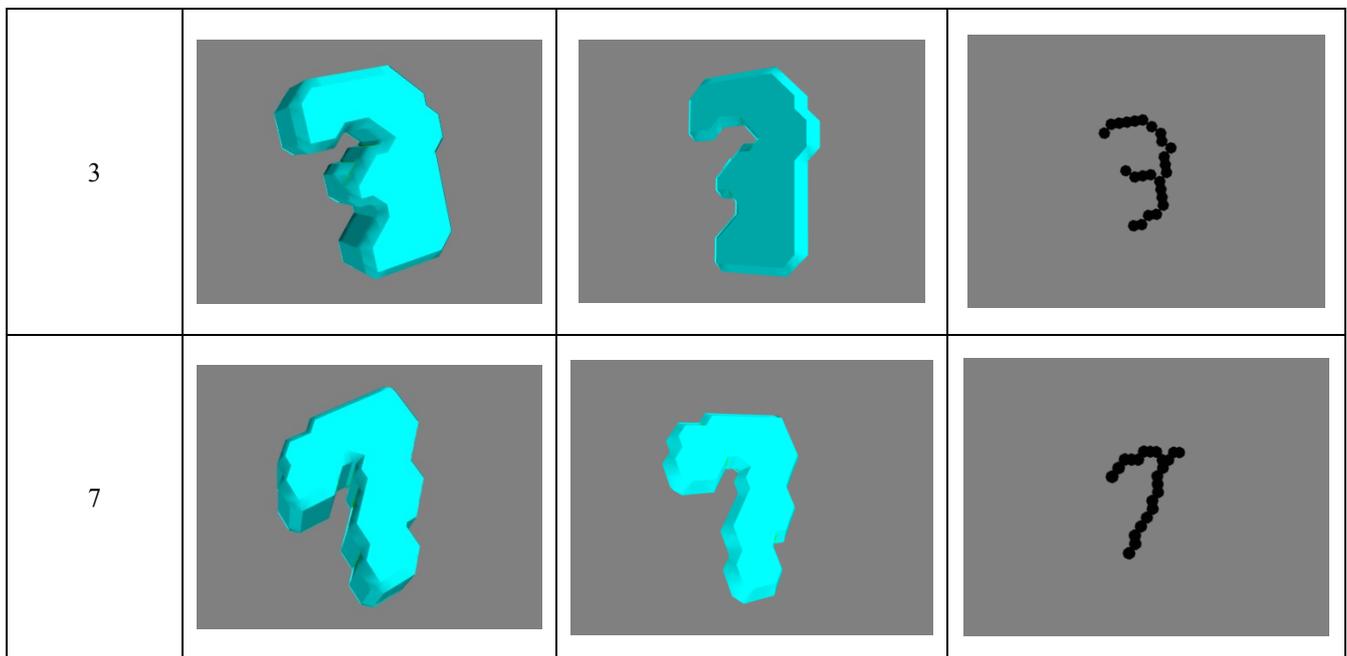

## VI. CONCLUSIONS AND FUTURE WORK

This paper proposed a new sequential thinning algorithm for multi-dimensional binary patterns. The method was tested on patterns of 2 and 3 dimensions and extensive experiments were performed to understand it's working. The proposed method is of a very "raw" form and is highly sensitive to noise.

The algorithm proved superior to two of the state-of-the-art methods discussed in terms of some quantitative measures and quality of the skeleton in the context of applications. The algorithm can also be modified to an extent depending on the needs of the application. The major drawback is its insensitivity to noise, which however lets it create skeletons that are highly accurate in structure.

Future work involves use of the proposed algorithm to detect boundary noise in binary patterns using skeleton models. Other work involves development of a preprocessing and/or postprocessing step to create better skeletons. Aside from these, the algorithm will also be tested in the context of different applications to understand its universality and limitations.

## REFERENCES

[1] Zhang, T. Y. and Suen, Ching Y. (1984). "A Fast Parallel Algorithms For Thinning Digital Patterns", Communication of the ACM, Vol 27, No. 3, Maret 1984, pp.236-239

[2] Guo, Z. and Hall, R.W. (1989). Parallel thinning with two-subiteration algorithms, Communications of the ACM 32(3): 359–373

[3] Jang, Ben K., and Roland T. Chin. "One-pass parallel thinning: analysis, properties, and quantitative evaluation." IEEE Transactions on Pattern Analysis and Machine Intelligence 14.11 (1992): 1129-1140

[4] Jang, B-K., and Roland T. Chin. "Analysis of thinning algorithms using mathematical morphology." IEEE Transactions on pattern analysis and machine intelligence 12.6 (1990): 541-551

[5] "K3M: A universal algorithm for image skeletonization and a review of thinning technique ", Int. J. Appl. Math. Comput. Sci., 2010, Vol. 20, No. 2, 317–335

[6] Gonzalez, R.C. and Woods, R.E. (2007). Digital Image Processing, 3rd Edition, Prentice Hall, Upper Saddle River, NJ.

[7] Ju, T., Baker, M.L. and Chiu, W. (2007). Computing a family of skeletons of volumetric models for shape description, Computer-Aided Design 39(5): 352–360

[8] Rosenfeld, A. (1975). A characterization of parallel thinning algorithms, Information and Control 29(3): 286–291

[9] Dyer, C. and Rosenfeld, A. (1979). Thinning algorithms for gray-scale pictures, IEEE Transactions on Pattern Analysis and Machine Intelligence 1(1): 88–89

[10] Pudney, Chris. "Distance-ordered homotopic thinning: a skeletonization algorithm for 3D digital images." *Computer Vision and Image Understanding* 72.3 (1998): 404-413.

[11] Ma, C. Min, and Milan Sonka. "A fully parallel 3D thinning algorithm and its applications." *Computer vision and image understanding* 64.3 (1996): 420-433.

[12] Tsao, Y. F., and King Sun Fu. "A parallel thinning algorithm for 3-D pictures." *Computer graphics and image processing* 17.4 (1981): 315-331.

[13] Pudney, Chris. "Distance-ordered homotopic thinning: a skeletonization algorithm for 3D digital images." *Computer Vision and Image Understanding* 72.3 (1998): 404-413.

[14] Lam, Louisa, Seong-Whan Lee, and Ching Y. Suen. "Thinning methodologies-a comprehensive survey." *IEEE Transactions on pattern analysis and machine intelligence* 14.9 (1992): 869-885.

[15] Lee, Ta-Chih, Rangasami L. Kashyap, and Chong-Nam Chu. "Building skeleton models via 3-D medial surface axis thinning algorithms." *CVGIP: Graphical Models and Image Processing* 56.6 (1994): 462-478.

[16] Palágyi, Kálmán, et al. "A sequential 3D thinning algorithm and its medical applications." *Biennial International Conference on Information Processing in Medical Imaging*. Springer, Berlin, Heidelberg, 2001.

[17] Saito, T. "A sequential thinning algorithm for three dimensional digital pictures using the Euclidean distance transformation." *Proc. 9th Scandinavian Conf. on Image Analysis, 1995. 6*. 1995.

[18] Palágyi, Kálmán, and Attila Kuba. "A parallel 3D 12-subiteration thinning algorithm." *Graphical Models and Image Processing* 61.4 (1999): 199-221.

[19] Xie, Wenjie, Robert P. Thompson, and Renato Perucchio. "A topology-preserving parallel 3D thinning algorithm for extracting the curve skeleton." *Pattern Recognition* 36.7 (2003): 1529-1544.

[20] Palágyi, Kálmán, and Attila Kuba. "A 3D 6-subiteration thinning algorithm for extracting medial lines." *Pattern Recognition Letters* 19.7 (1998): 613-627.

[21] Marti, U-V., and Horst Bunke. "The IAM-database: an English sentence database for offline handwriting recognition." *International Journal on Document Analysis and Recognition* 5.1 (2002): 39-46.

[22] https://www.kaggle.com/daavoo/3d-mnist/kernels.

[23] Bunke, Horst, and Tamás Varga. "Off-line Roman cursive handwriting recognition." *Digital Document Processing*. Springer London, 2007. 165-183.